\documentclass{sig-alternate}

\newcommand{\Real}{\mathop{\rm I\kern-.2emR}}
\newcommand{\DS}[0]{{\mathcal X}}
\newcommand{\OS}[0]{{\mathcal Z}}

\newcommand{\Neig}[0]{{\mathcal N}}

\newcommand{\setSol}[0]{\mathsf{\Sigma}}
\newcommand{\solSet}[0]{\mathtt{\sigma}}
\newcommand{\setN}[0]{\mathtt{N}}
\newcommand{\setFit}[0]{\mathtt{I}}
\newcommand{\IE}{\textit{i.e.} }

\usepackage{algorithmic,algorithm}

\begin{document}
\conferenceinfo{GECCO'11,} {July 12--16, 2011, Dublin, Ireland.}
\CopyrightYear{2011}
\crdata{978-1-4503-0557-0/11/07}
\clubpenalty=10000
\widowpenalty = 10000

%

\title{Set-based Multiobjective Fitness Landscapes:\\
A Preliminary Study}
%
%
%
%
%

\numberofauthors{3} 
%
\author{
%
%
\alignauthor S\'ebastien Verel\\
       \affaddr{Univ. Nice Sophia-Antipolis}\\
       \affaddr{INRIA Lille}\\
       \email{verel@i3s.unice.fr}
\alignauthor Arnaud Liefooghe
       \affaddr{Universit\'e Lille 1}\\
       \affaddr{LIFL -- CNRS -- INRIA Lille}\\
       \email{arnaud.liefooghe@lifl.fr}
\alignauthor Clarisse Dhaenens
       \affaddr{Universit\'e Lille 1}\\
       \affaddr{LIFL -- CNRS -- INRIA Lille}\\
       \email{clarisse.dhaenens@lifl.fr}
}
\date{9 February 2011}

\maketitle

\begin{abstract}
Fitness landscape analysis aims to understand the geometry of a given optimization problem in order to design more efficient search algorithms. However, there is a very little knowledge on the landscape of multiobjective problems. In this work, following a recent proposal by \textit{Zitzler et al.~(2010)}, we consider multiobjective optimization as a set problem. Then, we give a general definition of set-based multiobjective fitness landscapes. An experimental set-based fitness landscape analysis is conducted on the multiobjective $NK$-landscapes with objective correlation. The aim is to adapt and to enhance the comprehensive design of set-based multiobjective search approaches, motivated by an \textit{a priori} analysis of the corresponding set problem properties.
\end{abstract}

\category{F.2.m}{Analysis of Algorithms and Problem Complexity}{Miscellaneous}

\terms{Algorithms}

\keywords{Fitness landscapes, Multiobjective optimization, Set-based multiobjective search}

\section{Introduction}
There exists a large amount of literature about multiobjective optimization in general,
and about the identification or the approximation of the Pareto optimal set in particular.
In the latter case, evolutionary multiobjective optimization~(EMO) techniques have received a growing interest since the late 1980s.
The overall goal is generally to identify a set of good-quality solutions (ideally the whole or a `representative' subset of the Pareto optimal set).
As a consequence, recent advances in the field explicitly formulate the goal of multiobjective optimization as a set problem:
the search space is made of sets of solutions (and not single solutions)~\cite{zitzler2010}.
However, to date,
the impact of the main problem-related properties on the behavior and the performance of set-based multiobjective search approaches
is still far from being well-understood.

Up to now, the definition of multiobjective fitness landscapes (moFiL) has been mainly restricted upon two different levels:
the properties of the Pareto optimal set on the one hand, and of the search space properties at the solution-level on the other hand.
With respect to the Pareto optimal set,
problem-related properties are known to largely affect the structure of Pareto optimal solutions \cite{Mote1991},
and then the behavior of search algorithms \cite{paquete2006}.
With respect to the solution-level,
Knowles and Corne \cite{knowles2002} lead a landscape analysis
on the multiobjective quadratic assignment problem with a rough objective correlation.
The transposition of standard tools from fitness landscape analysis to multiobjective optimization are discussed by Garrett~\cite{garrett2007}, 
together with a study on fitness-distance correlation.
In another study, a moFiL is regarded as a neutral landscape, 
and divided into different fronts with the same dominance rank~\cite{Garrett09}.
In previous works on multiobjective $NK$-landscapes~\cite{aguirre2007},
enumerable moFiL are studied according to the number of fronts, the number of solutions on each front,
the probability to pass from one front to another, and the hypervolume-value of the Pareto optimal set. 

These  types of moFiL lead to rather poor tools to describe 
the dynamics of population-based multiobjective search algorithms.
We here propose to define a third type of moFiL, dealing with the \emph{search space properties at the set-level}.
The contributions of this work are summarized below.
\begin{itemize}
\item[($i$)] A definition of \emph{set-based multiobjective fitness landscapes} is given,
based on a search space made of solution-sets, a neighborhood relation between solution-sets, and an indicator-based fitness function. 
\item[($ii$)] An experimental analysis is conducted
in order to study standard tools from single-objective fitness landscapes (ruggedness and multimodality)
in the context of set-based multiobjective search.
We study the influence of the main problem-related properties 
and of the solution-set size on multiobjective $NK$-landscapes.
\end{itemize}
The reminder of the paper is organized as follows.
Section~\ref{sec:background} deals with fitness landscapes, multiobjective optimization, and set-based multiobjective search. 
In section~\ref{sec:setbased}, we give a definition of set-based multiobjective fitness landscapes, illustrated with multiple examples.
Experimental results are given in Section~\ref{sec:exp};
and the last section concludes the paper.

\section{Preliminaries}
\label{sec:background}

\subsection{Fitness Landscapes}
\label{sec:monoFil}

In single-objective optimization,
the notion of fitness landscape (FiL) has been introduced to study the topology of a problem \cite{jones95}.
A FiL can be defined by the triplet~$(S, \Neig, f)$ such that:
$S$ is a set of admissible solutions ({\itshape i.e.} the search space);
$\Neig: S \rightarrow 2^S$ is a \emph{neighborhood relation},
\IE a function that assigns a set of solutions $\Neig(s) \subset S$ to any solution $s \in S$,
the set $\Neig(s)$ is called the \emph{neighborhood} of $s$, and a solution $s^\prime \in \Neig(s)$ is called a \emph{neighbor} of $s$;
$f: S \longrightarrow \Real$ is a fitness function that can be pictured as the `\textit{height}' of the corresponding solutions,
here assumed to be maximized.
A \emph{local optimum} is a solution $s^{\star} \in S$ such that $\forall s \in \Neig(s^{\star})$, $f(s) \leq f(s^{\star})$.
The ability of search algorithms is related to the number of local optima, 
and to their distribution over the landscape \cite{Merz2004}.
\emph{Global optima} are defined as the absolute maxima in the whole search space $S$.
Other landscape features, such as basins, barriers, or neutrality can also be defined \cite{stadler-02}. 
For the sake of self-containedness,
several notions that will be used later in the paper are defined below.

A \emph{walk} on the landscape from $s$ to $s^\prime \in S$ is a sequence
$(s_0, s_1, \ldots, s_m)$ of solutions from the search space
such that $s_0=s$, $s_m = s^\prime$ and $s_i \in \Neig(s_{i-1})$ $\forall i \in \{1,\dots,m\}$.
For instance, the walk is said to be random if solutions are chosen with a uniform probability from the neighborhood. 
It can also be obtained through the repeated application of a `move' operator defined on the landscape,
such as a random mutation or a deterministic hill-climbing. 

Given a walk $( s_t, s_{t+1}, \ldots  )$,
the \emph{autocorrelation function}~\cite{WEI:90} ($\rho$) of a fitness function $f$ is the autocorrelation
function of time series $( f(s_t), f(s_{t+1}), \ldots )$:
$$
\rho(k) = \frac{E[f(s_t) f(s_{t+k})] - E[f(s_t)]E[f(s_{t+k})]}{var(f(s_t))}
$$
where $E[f(s_t)]$ and $var(f(s_t))$ are the expected value and the variance of $f(s_t)$, respectively.
Estimates $r(k)$ of autocorrelation coefficients $\rho(k)$ can be calculated with a time
series $( s_1, s_2, \ldots, s_{L} )$ of length $L$:
$$
r(k) = \frac{\sum_{j=1}^{L-k}(f(s_j) - \bar{f}) (f(s_{j+k}) - \bar{f})}{\sum_{j=1}^{L}(f(s_j) - \bar{f})^2}
$$
where $\bar{f} = \frac{1}{L} \sum_{j=1}^{L} f(s_j)$, and $L >> 0$.
A random walk is representative of the landscape when it is statistically isotropic.
In such a case, whatever the starting point and the neighbors selected during the walks,
estimates of $r(n)$ are nearly the same.
The estimation error diminishes with the length of the walk.
The \emph{autocorrelation length} $\tau$ measures how the autocorrelation function decreases.
This summarizes the ruggedness of the landscape: the larger the correlation length, the smoother the landscape.
Weinberger's definition $\tau = - \frac{1}{ln(\rho(1))}$ makes the assumption that
the autocorrelation function decreases exponentially \cite{WEI:90}. 

The length of \emph{adaptive walks}, performed with a hill-climber,
is an estimator of the diameter of the local optima basins of attraction.
The larger the length, the larger the basin size.
This allows to estimate the number of local optima when the whole search space cannot be enumerated exhaustively.

\subsection{Multiobjective Optimization}
A multiobjective optimization problem can be defined by
a set of $M \geq 2$ objective functions $F=(f_1, f_2,\dots, f_M)$,
and a set $\DS$ of feasible solutions in the \emph{decision space}.
In the combinatorial case, $\DS$ is a discrete set.
Let $\OS  =  F(\DS) \subseteq \Real^M$ be the set of feasible outcome vectors in the  \emph{objective space}.  
In a maximization context, a solution $x^\prime \in \DS$ is dominated by a solution $x \in \DS$, denoted by $x^\prime \prec x$,
iff $\forall i \in \{1,2,\dots,M\}$, $f_i(x^\prime) \leq f_i(x)$ and $\exists j \in \{1,2,\dots,M\} $ such that $f_j(x^\prime) < f_j(x)$.
A solution $x^\star \in \DS$ is said to be \emph{Pareto optimal} (or \emph{efficient}, \emph{non-dominated}), 
if there does not exist any other solution $x \in \DS$ such that $x^\star \prec x$.
The set of all Pareto optimal solutions is called the \emph{Pareto optimal set} (or the \emph{efficient set}).
Its mapping in the objective space is called the \emph{Pareto front}.
A common approach is to identify a minimal complete Pareto optimal set,
for which each point of the Pareto front corresponds to a single Pareto optimal solution.

However, generating the entire Pareto optimal set is often infeasible for two main reasons: 
($i$) the number of Pareto optimal solutions is typically exponential in the size of the problem instance,
and ($ii$) deciding if a feasible solution belongs to the Pareto optimal set is often NP-complete.
Therefore, the overall goal is often to identify a good \emph{Pareto set approximation}.
To this end, evolutionary algorithms  have received a growing interest since the late eighties.

\subsection{Set-based Multiobjective Search}
\label{sec-setbased}

Recently, approximating the Pareto optimal set has been explicitly stated as a \emph{set problem} \cite{zitzler2010}.
In that sense, most existing EMO algorithms can be seen as hill-climbers performing on sets.
Let us define the search space $\setSol \subset 2^\DS$ by a set of feasible sets of solutions (and not single solutions).
An element $\solSet \in \setSol$ is denoted as a \emph{solution-set}.
Usually, a maximum cardinality is imposed: $|\solSet| \leq \mu$ for all $\solSet \in \setSol$.
Different interpretations of what is a good Pareto set approximation are possible, 
and the definition of approximation quality strongly depends on the decision-maker preferences.
A \emph{set preference relation} is then usually induced over~$\setSol$, like the Pareto dominance relation extended to solution-sets.

We here assume that the set preference relation is explicitly given in terms of a quality indicator $\setFit : \setSol \rightarrow \Real$.
One of them is the \emph{hypervolume indicator} $\setFit_H$ \cite{zitzler2003}, that is to be maximized.
It gives the portion of the objective space enclosed by a solution-set $\solSet \in \setSol$ and a reference point $z^\star \in \OS$.
The hypervolume indicator is one of the most commonly used indicator, due to several interesting properties \cite{zitzler2007}.
In particular, this is the only indicator that is dominance preserving,
\IE $\forall \solSet, \solSet^\prime \in \setSol $ such that $\solSet^\prime$ is dominated by $\solSet$:
$\setFit_H(\solSet) \geq \setFit_H(\solSet^\prime)$.
Many recent search algorithms are based on the hypervolume indicator,
but most of them operates at the solution-level \cite{beume2007,zitzler2004}, with the exception of~\cite{bader2008}.
The goal of a hypervolume-based search is then to find a solution-set $\solSet \in \setSol$ that maximizes the indicator value:
\begin{equation}
\arg \max_{\solSet \in \setSol}\ \setFit_H(\solSet)
\label{eq:goal}
\end{equation}
Let us note that a minimal solution-set maximizing $\setFit_H$ is a subset of the Pareto optimal set.
Therefore, $\setFit_H$ can be seen as a function that assigns, to each solution-set, 
a scalar value reflecting its quality according to the goal formulated in (\ref{eq:goal}),
\IE a \emph{fitness function} defined over sets.

\section{Set-based Fitness Landscapes}
\label{sec:setbased}

\subsection{Definition}

Like in single-objective optimization, a multiobjective fitness landscape (moFiL) requires a proper definition of
($i$)~a search space, 
($ii$)~a neighborhood operator, and
($iii$)~a fitness function. 
From a multiobjective perspective, several remarks and criticisms can be stated from previous attempts made in the past in defining a moFiL.
First, the output of a multiobjective search algorithm is a solution-set,
and not a single solution like in the single-objective case.
Moreover,
multiobjective search approaches in general manipulate either a population of solutions, or an archive of mutually non-dominated solutions.
Both can be viewed as solution-sets.
As a consequence, following the work of Zitzler et al.~\cite{zitzler2010},
identifying multiple tradeoff solutions by means of a Pareto set approximation
can explicitly be stated as a set problem (see Section \ref{sec-setbased}). 
The search space of a multiobjective optimization problem is here assumed to be constituted of a set of feasible solution-sets.
Second, considering a partial order only to analyze a FiL
does not allow to measure interesting fitness landscape features dealing with the ruggedness and the evolvability (among others).
This is the reason why the Pareto dominance relation (or a slight modification of it) is generally not satisfying enough to define a moFiL.
Quality indicators as defined in~\cite{zitzler2003} allow to overcome such a limitation by introducing a complete order between solution-sets,
and by quantifying their respective quality with respect to the indicator being used.
Last, in their proposal on set-based multiobjective search,
Zitzler et al.~\cite{zitzler2010} do not define any set-based neighborhood operator,
then restricting the application of their approach to some `\textit{random set mutation}', or `\textit{heuristic set mutation}'.
However, defining a neighborhood structure on solution-sets allows to distinguish between the properties of the search space,
and the heuristics used to explore solution-set's neighborhood.
This is also through this definition that are located the main differences in the dynamics of set-based multiobjective search algorithms.

In this work, we propose the definition of a moFiL in terms of set-based multiobjective search by means of an indicator-based fitness function.
\\

\noindent
\fbox{
\begin{minipage}{0.94\columnwidth}

A \textit{set-based multiobjective fitness landscape} is defined as a triplet~$(\setSol, \setN, \setFit)$
such that:
\begin{itemize} 
\item
$\setSol \subset 2^\DS$ is a set of feasible solution-sets (where $\DS$ is the set of feasible solutions);
\item
$\setN : \setSol \rightarrow 2^{\setSol}$ is a neighborhood relation between solution-sets;
\item
$\setFit : \setSol \rightarrow \Real$ is a unary quality indicator,
\IE a fitness function measuring the quality of solution-sets.

\end{itemize}

\end{minipage}
}
\null\\

\noindent
$\setSol$, $\setN$, and $\setFit$ still need to be defined for the problem at hand.
But this is also the case in single-objective optimization, except that they are here defined at the set-level.

Algorithm~\ref{algo} gives a general class of algorithms that set-based moFiL are able to compare.
For sure, most existing multiobjective search algorithms can be formulated as instances of this general methodology.
\begin{algorithm}[t]
\caption{Set-based Neighborhood Search Algorithm}
\begin{algorithmic}
\STATE start with a solution-set $\solSet \in \setSol$ 
\STATE evaluate $\solSet$ with respect to $\setFit$
\REPEAT
   \STATE select $\solSet^\prime \in \setN(\solSet)$
   \STATE evaluate $\solSet^\prime$ with respect to $\setFit$
   \IF{accept($\solSet$,$\solSet^\prime$)}
      \STATE $\solSet \leftarrow \solSet^\prime$
   \ENDIF
\UNTIL ({continue($\solSet$)})
\STATE \textbf{return} non-dominated solutions of $\solSet$
\end{algorithmic}
\label{algo}
\end{algorithm}

\subsection{Illustrative Examples of Set-based moFiL}

Different set-level search spaces can be considered according to the problem and the algorithm under study.
Several examples are given below.
\begin{itemize}
\item 
The search space of population-based approaches can be defined as
$\setSol = \{ \solSet \in 2^{\DS}~:~|\solSet| = \mu \}$, where $\mu$ is the population size.
\item
The search space of approaches using a bounded archive can be defined as
$\setSol = \{ \solSet \in 2^{\DS}~:~|\solSet| \leq \mu \}$, where $\mu$ is the maximal size of the archive.
\item
The search space of a number of existing dominance-based approaches,
where solution-sets of mutually non-dominated solutions only are considered,
can be defined as $\setSol = \{ \solSet \in 2^{\DS}~:~\forall s, s^\prime \in \solSet , s \not \prec  s^\prime\}$.
\item
A search space with the two previous restrictive conditions can also be considered,
\IE $\setSol = \{ \solSet \in 2^{\DS}~:~|\solSet| \leq \mu \text{ and } \forall s,s^\prime \in \solSet , s \not \prec  s^\prime \}$.
\item
A search space without any restriction is $\setSol = 2^{\DS}$.
\end{itemize}
%
Next,
the neighborhood structure has to reflect the way the search space is explored by a class of search algorithms.
In the general case, the definition of neighborhood is based either on a distance,
or more often on the variation operator(s) handled by the algorithm under study.
Roughly speaking, at the set-level, the neighbors of a solution-set can for instance be obtained by
($i$)~replacing a solution from the set,
($ii$)~inserting a solution to the set, or
($iii$)~deleting a solution from the set.
In order to give more precise examples of set-level neighborhood operators,
let us consider an arbitrary non-empty solution-set $\solSet \in \setSol$,
an arbitrary non-empty neighboring solution-set $\solSet^\prime = \setN(\solSet)$,
and an arbitrary neighboring solution $s^\prime \in \Neig(s)$ with $s \in \solSet$.
Possible set-level neighborhood operators are discussed below.
\begin{itemize}
\item
When replacing a solution from the set, a neighboring solution-set can be defined as
$\solSet^\prime = \solSet \cup \{ s^\prime \} \setminus \{ s^{\prime\prime} \}$ such that $s^{\prime\prime} \in \solSet$.
The size of this replacement set-level neighborhood is at most $| \solSet | \cdot \sum_{s \in \solSet} |\Neig(s)|$.
\end{itemize}
In such a case, a possible neighborhood exploration strategy is to find the tuple $(s^\prime,s^{\prime\prime})$,
with $s^{\prime\prime} \in \solSet$,
such that $\setFit(\solSet \cup \{ s^\prime \} \setminus \{ s^{\prime\prime} \})$ is maximal.
However, most existing EMO methodologies generally separate the fact of inserting a solution to the set,
and deleting a solution from the set into two different phases.
\begin{itemize}
\item
When inserting a new solution to the set, a neighboring solution-set can be defined as
$\solSet^\prime = \solSet \cup \{ s^\prime \}$.
The size of this insertion neighborhood is at most $\sum_{s \in \solSet} |\Neig(s)|$.
\item
When deleting a solution from the set, a neighboring solution-set can be defined as
$\solSet^\prime = \solSet \setminus \{ s \}$ where $s \in \solSet$.
The size of this deletion neighborhood is $| \solSet |$.
\end{itemize}
%
The set-level neighborhood operators can be applied multiple times in order to define large-size neighborhood operators,
where several solutions can differ in a neighboring solution-set.
A neighboring solution-set must always correspond to an element of the given search space.
As a consequence, when the solution-sets are somehow bounded in size, 
the neighborhood must be restricted using a (partial) dominance relation, or a limited-size set.
Of course, the definition of a set-level neighborhood relations are not limited to the use of
a solution-level neighboring operator $\Neig$.
For instance, a set-level neighborhood relation can consider a random solution,
or a solution produced by applying a recombination operator to pairs of solutions in the solution-set, and so on.
Anyway, all those set-level neighborhood operators are just few examples,
and like in single-objective optimization,
one has to define the neighborhood relation according to the (set) problem and the algorithm under study.

At last, the fitness function defined for set-based moFiL is given in terms of a quality indicator.
Several studies are devoted to theoretical properties of multiobjective quality indicators.
Fore more details, the reader is referred to \cite{zitzler2003}.

\subsection{Discussion}

In the general case, two typical uses of FiL analysis can be conducted.
First, such a study can allow to compare the difficulties, in terms of FiL features,
associated with different search problems.
Given a search algorithm and two different optimization problems,
the corresponding FiL are defined (\IE the search space, the neighborhood relation, and the fitness function).
Then, the difficulties can be compared between both FiL according to measures dealing, for instance,
with the number of local optima, their distribution, the ruggedness, the evolvability, and so on.
Second, another possibility of FiL analysis is the off-line tuning or design of search approaches.
Once again, given a search problem and different possible component design or parameter setting,
the corresponding landscapes are defined.
Then, according to the FiL measures, the most promising search algorithm components can be chosen \textit{a priori}.
In the context of set-based multiobjective search, a comparison of two set-based moFiL
can be compared with each other in terms of FiL measures.
They can be defined, for instance, by two different neighborhood operators, two different fitness functions or two different search space definitions,
In the following, we conduct an empirical study on the comparison of difficulty of multiobjective optimization problems.

\section{Experimental Analysis}
\label{sec:exp}

\subsection{$\rho MNK$-Landscapes}
\label{sec:rhomnk}

In the single-objective case, 
the family of $NK$-landscapes constitutes an interesting model to study the influence of non-linearity on the number of local optima.
In this section, we present the $\rho MNK$-landscapes proposed in \cite{verel2011b}.
Four parameters are required to define a $\rho MNK$-landscape:
the problem size~$N$,
the number of epistatic links~$K$, 
the number of objectives~$M$,
and
the objective correlation coefficient~$\rho$.

The family of $NK$-landscapes is a problem-independent model used for constructing multimodal landscapes \cite{kauffman93}.
Parameter $N$ refers to the number of bits in the decision space (\IE the string length)
and $K$ to the number of bits that influence a particular bit from the string (the epistatic interactions).
By increasing the value of $K$ from 0 to $(N-1)$, $NK$-landscapes can be gradually tuned from smooth to rugged.
The fitness function (to be maximized) of a $NK$-landscape $f_{NK}: \lbrace 0, 1 \rbrace^{N} \rightarrow [0,1)$
is then defined on binary strings of size $N$. 
An `atom' with a fixed epistasis level is represented by a fitness component $f_i: \lbrace 0, 1 \rbrace^{K+1} \rightarrow [0,1)$
associated with each bit~$i \in N$.
Its value depends on the allele at bit~$i$ and also on the alleles at $K$ other bit positions ($K$ must fall between $0$ and $N - 1$).
In other words, the parameter $K$ tunes the degree of non-linearity (epistasis).
The fitness $f_{NK}(x)$ of a solution $x \in \lbrace 0, 1 \rbrace^{N}$ corresponds to the mean value of its $N$ fitness components $f_i$:
$f_{NK}(x) = \frac{1}{N} \sum_{i=1}^{N} f_i(x_i, x_{i_1}, \ldots, x_{i_K})$,
where $\lbrace i_1, \ldots, i_{K} \rbrace \subset \lbrace 1, \ldots, i-1, i+1, \ldots, N \rbrace$.
In this work, we set the $K$ bits randomly.
Each fitness component $f_i$ is specified by extension,
\IE a number $y^i_{x_i, x_{i_1}, \ldots, x_{i_K}}$ from $[0, 1)$ is associated with each element
$(x_i, x_{i_1}, \ldots, x_{i_K})$ from $\lbrace 0, 1 \rbrace^{K+1}$.
Those numbers are uniformly distributed in the range $[0, 1)$. 
As a consequence, it is very unlikely that  the same fitness value is assigned to two different solutions.

A multiobjective variant of $NK$-landscapes (namely $MNK$-landscapes)
has been defined with a set of $M$ independent fitness functions \cite{aguirre2007}.
The same epistasis degree $K_{m}=K$ is used for all the objectives.
Each fitness component $f_{m, i}$ is specified by extension
with the numbers $y^{m,i}_{x_i, x_{i_{m, 1}}, \ldots, x_{i_{m, K_m}}}$.
In the original $MNK$-landscapes, these numbers are defined randomly and independently. 
An approach for designing $MNK$-landscapes with correlated objective functions has been recently proposed in~\cite{verel2011b}.
First, let us define the $CMNK$-landscapes, where the epistasis structure is identical for all the objective functions:
$\forall m \in \{1,\dots,M\}$,  $K_{m}=K$
and
$\forall m \in \{1,\dots,M\}$, $\forall j \in \{1,\dots,K\}$, $i_{m, j} = i_{j}$.
However, the fitness components are not defined independently.
The numbers $(y^{1,i}_{x_i, x_{i_1}, \ldots, x_{i_{K}}}, \ldots, y^{M,i}_{x_i, x_{i_1}, \ldots, x_{i_{K}}})$
follow a multivariate uniform law of dimension~$M$,
defined by a correlation matrix~$C$.
Thus, the $y$'s follow a multidimensional law with uniform marginals 
and the correlations between $y^{m, i}_{\ldots}$s are defined by the matrix $C$.
The construction of $CMNK$-landscapes defines correlation between the $y$'s but not directly between the objectives. 
In \cite{verel2011b}, it is proven by algebra that the correlation between objectives is tuned by the matrix $C$:
$E(cor(f_{n}, f_{p})) = c_{np}$.
In the $\rho MNK$-landscapes,
the correlation matrix $C_{\rho}=(c_{np})$ is assumed to have the same correlation between all pairs of objectives:
$c_{nn} = 1$ for all $n$, and $c_{np} = \rho$ for all $n \not= p$.
Of course, for obvious reasons, it is not possible to have the matrix $C_{\rho}$ for all $\rho$ values in $[-1,1]$:
$\rho$ must be greater than $\frac{-1}{M-1}$, see \cite{verel2011b}.
In $\rho MNK$-landscape, the parameter~$\rho$ allows to tune very precisely the correlation between all pairs of objectives.

In the following, we conduct an empirical study of the influence of
the problem dimension, the non-linearity~(epistasis), the number of objective functions and the objective correlation
on some properties of set-based moFiL.

\subsection{Experimental Design}

In order to minimize the influence of the random creation of landscapes, 
we considered $30$ different and independent instances for each parameter combinations: $\rho$, $M$, and $K$.
The measures reported are the average over these 30 instances. 
The parameters under investigation in this study are given in Table~\ref{tab:param}.
\begin{table}
\caption{Parameters used in the paper.} 
\begin{center}
\begin{small}
\begin{tabular}{c|l}
Parameter & Values \\
\hline
$N$ & $\{64\}$\\
$M$ & $\{2, 3, 5 \}$ \\
$K$ & $\{2, 4, 6, 8, 10 \}$\\
$\rho$ &  $\{ -0.9, -0.7, -0.4, -0.2, 0.0, 0.2, 0.4, 0.7, 0.9 \}$\\
&  such that $\rho \geq -1 / (M-1)$ \\
\end{tabular}
\end{small}
\end{center}
\label{tab:param}
\end{table}
%
We analyze the multiobjective $\rho MNK$-landscapes according to set-based search algorithms that manipulate a fixed-size solution-set.
The goal is to show the link between the geometry of the set-based moFiL
and the features that make a search algorithm efficient for the corresponding problem.
Previous results indicate that the problem is getting more complex when the non-linearity and the degree of conflict between the objectives are high
\cite{aguirre2007,verel2011b}.
Feasible solutions are bit strings of size $N$: $\DS = \{ 0, 1 \}^N$,
and the set-level search space is the set of solution-sets of size~$\mu$.
The set-level neighborhood relation consists of the replacement neighborhood as defined in Section~\ref{sec:setbased}.
It does not change the solution-set size and uses a bit-flip solution-level neighborhood operator.
In this work, we do not consider the possible insertion or deletion of solutions from the solution-set.
Hence, two solution-sets are neighbors if they have the same size, and if they differ by one solution only. 
It is also required that the corresponding solutions are neighbors according to the one bit-flip neighborhood operator:
$\solSet^\prime \in \setN(\solSet)$ iff 
$\exists s \in \solSet $, $\exists s^\prime \in \DS$ 
such that $d_{Hamming} (s^\prime, s) = 1$ and $\solSet^\prime = \solSet \setminus \{ s \} \cup \{ s^\prime \}$.
The maximal size of this neighborhood relation is then $(| \solSet | \cdot N)$.
The set-level fitness function is based on the hypervolume indicator \cite{zitzler2003}.
Given that the objective functions of $\rho MNK$-landscapes, defined in $[ 0 , 1 ]$, are to be maximized,
the reference point required by the hypervolume calculation is set to~$0^{M}$.

\subsection{Ruggedness}

\begin{figure}
\begin{center}
\includegraphics[width=0.345\textwidth]{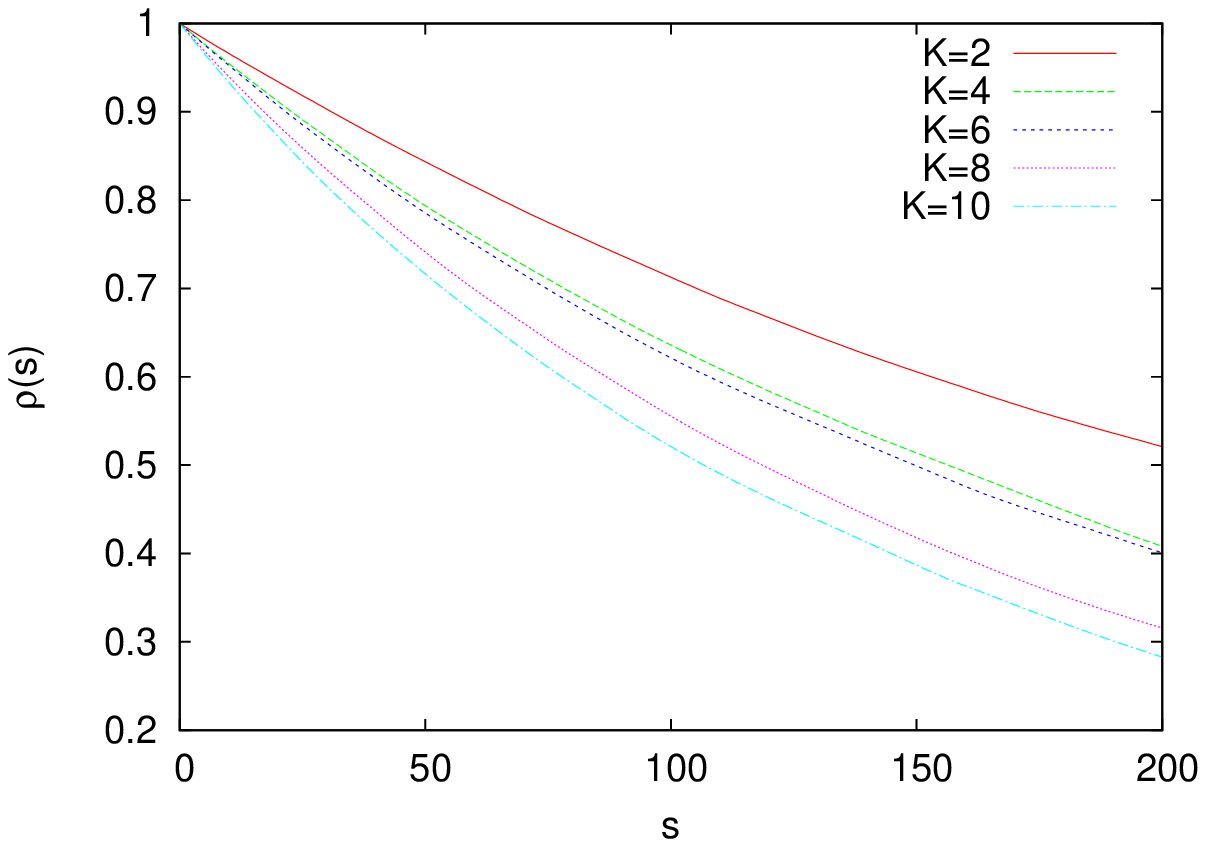} \\
\includegraphics[width=0.345\textwidth]{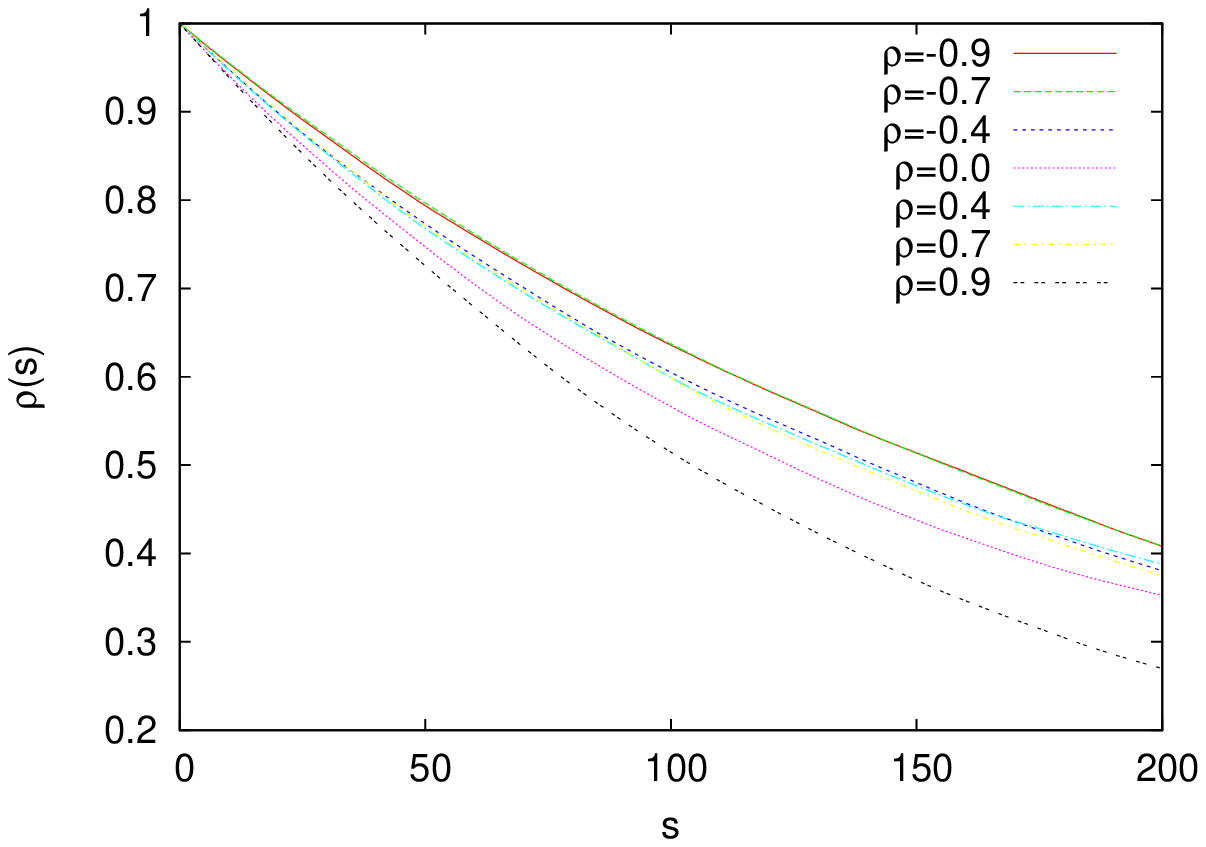} 
\caption{Autocorrelation functions
according to parameter $K$ (top $\rho=-0.2$),
and to parameter $\rho$ (bottom $K=2$).
The number of objectives is $M=2$.
\label{fig:autocorFunction}}
\end{center}
\end{figure}
%
\begin{figure*}[p]
\begin{center}
\begin{tabular}{cc}
\includegraphics[width=0.34\textwidth]{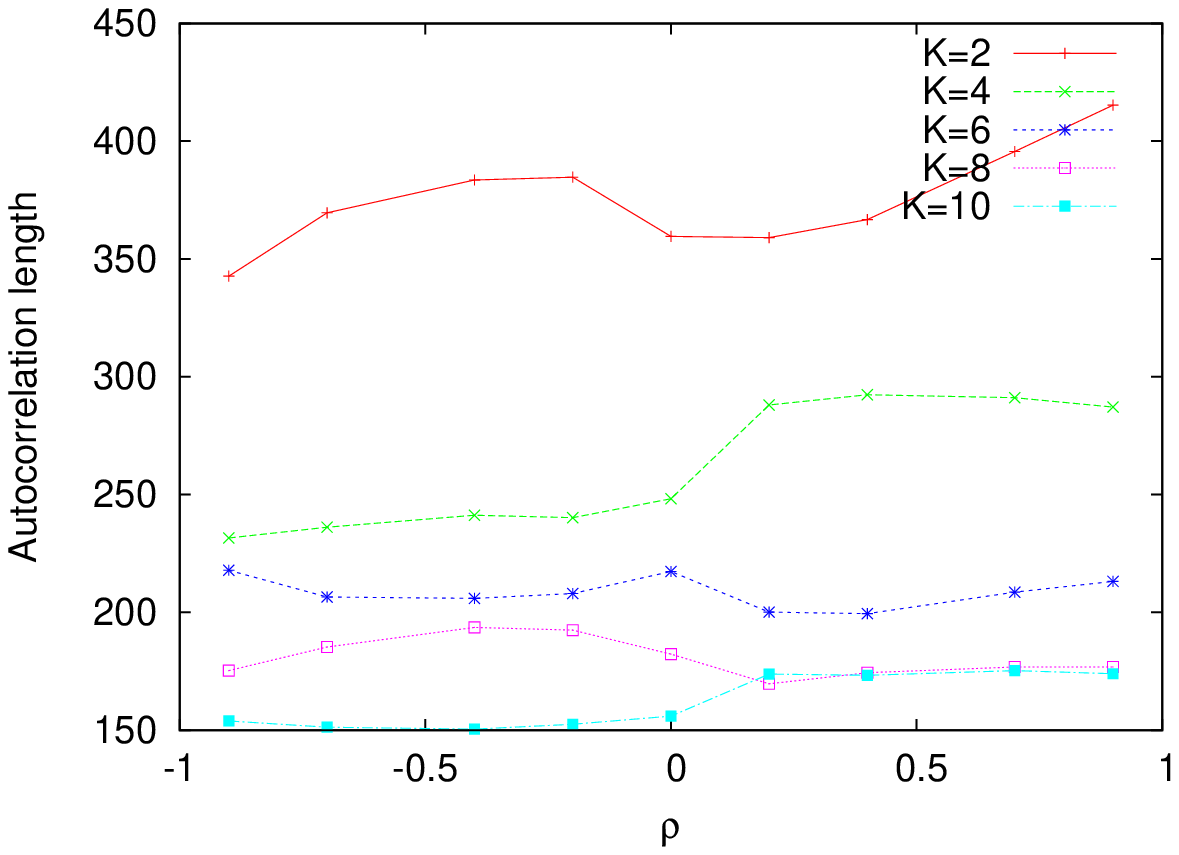} & \includegraphics[width=0.34\textwidth]{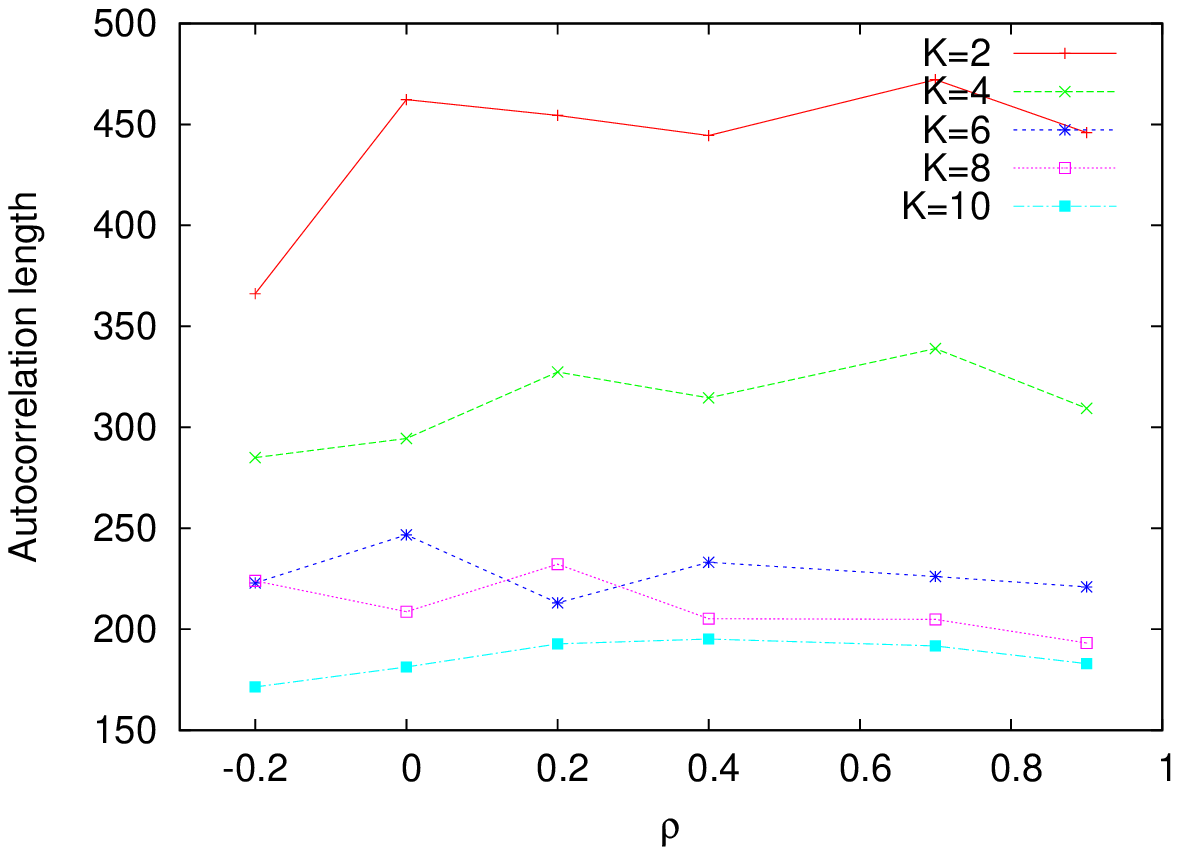} \\
\includegraphics[width=0.34\textwidth]{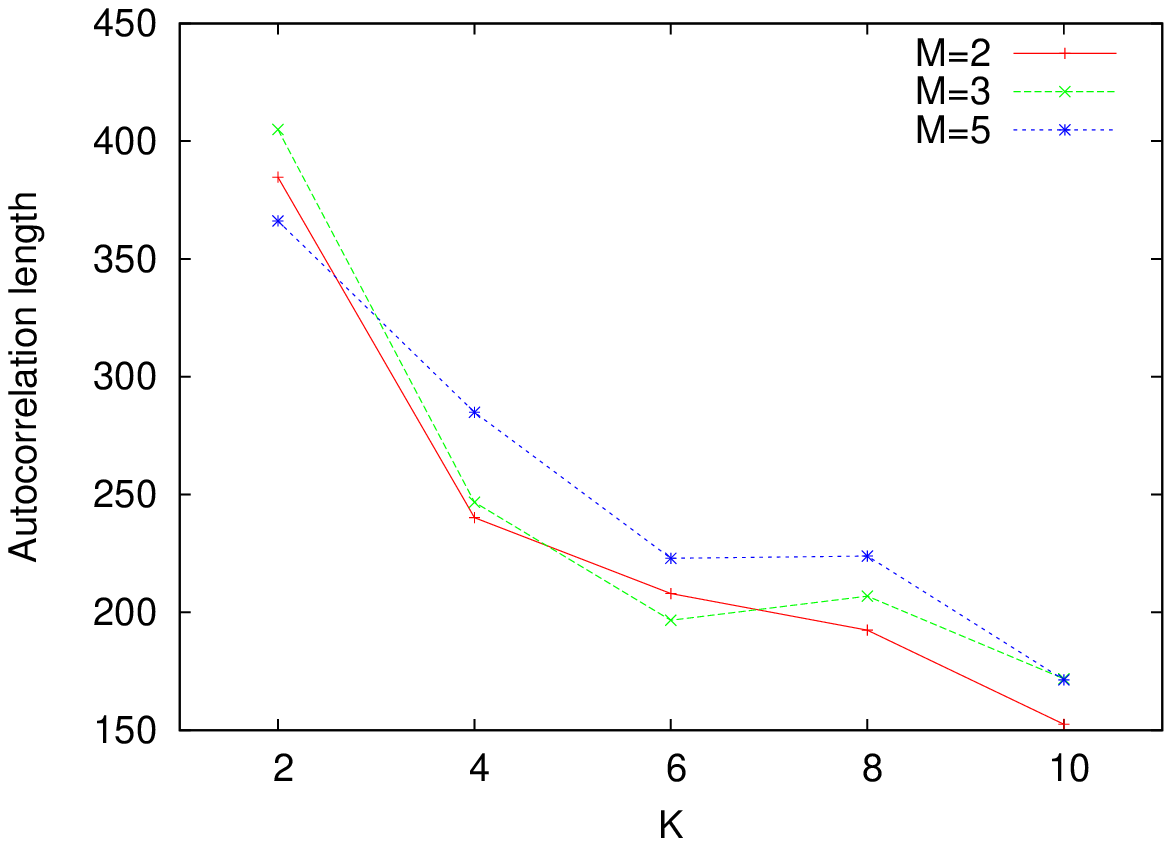} & \includegraphics[width=0.34\textwidth]{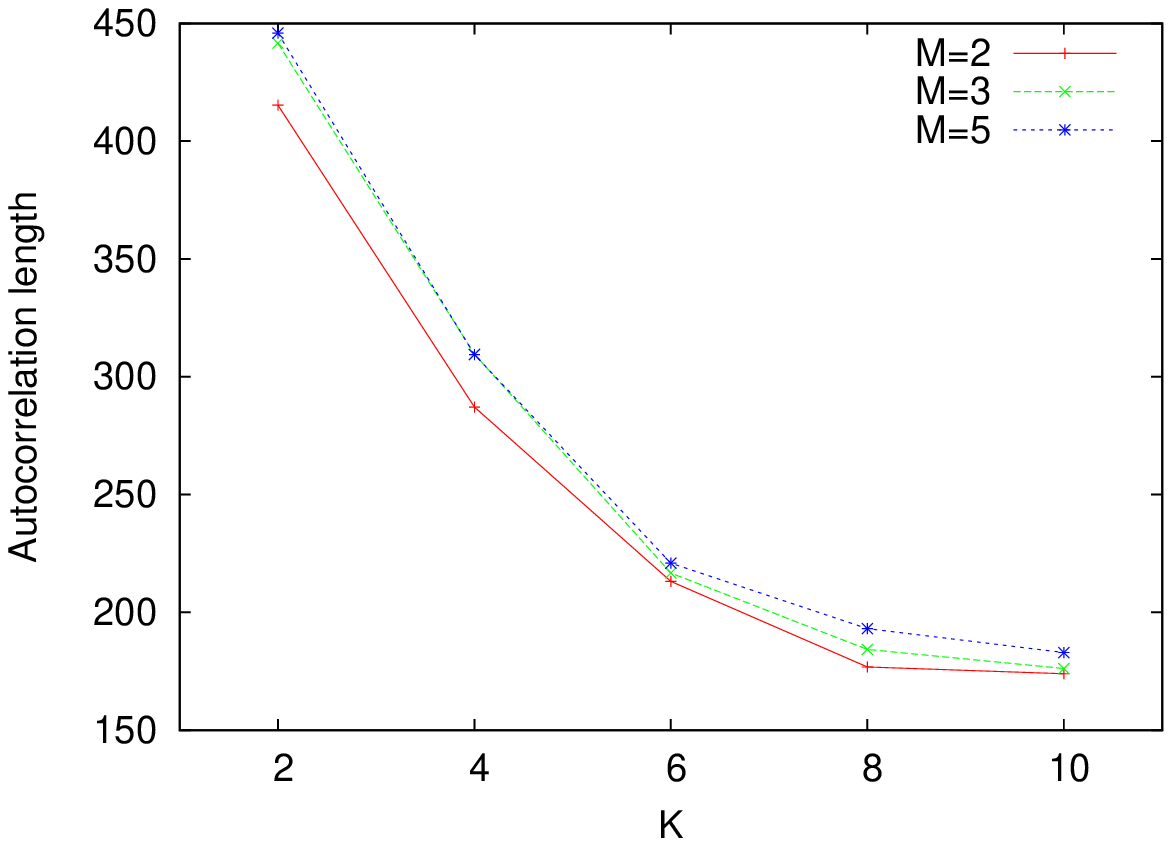} \\
\end{tabular}
\caption{Average value of the autocorrelation length
according to parameter $\rho$ (top left $M=2$, right $M=5$),
and to parameter $K$ 
(bottom left $\rho=-0.2$, right $\rho=0.9$).
The solution-set size is $\mu=100$.
\label{fig:autocorLength}}
\end{center}
\end{figure*}

The ruggedness of a multiobjective problem is here measured
in terms of the autocorrelation of the hypervolume along a random walk. 
The starting solution-set of the walk is initialized with $\mu=100$ random solutions.
At each step of the random walk, a random neighboring solution-set replaces the current one.
The length of the random walk is set to $5.10^3$.
Figure \ref{fig:autocorFunction} shows the autocorrelation functions for an objective space dimension $M=2$
with respect to the non-linearity degree $K$, and to the objective correlation $\rho$.
The functions all decrease slowly with the step lag.
The hypervolume correlation between random neighboring solution-sets is high.
Figure \ref{fig:autocorLength} shows the autocorrelation length 
according to parameter $K,M$ and $\rho$.
The correlation values are very high.
As a comparison, the autocorrelation length of single-objective $NK$-landscapes is $-1 / \log(1- \frac{K+1}{N})$,
which gives the length $20.8$ for $N=64$ and $K=2$ \cite{stadler-02}.
The correlation between neighboring solutions with respect to each objective function
impacts the correlation between neighboring solution-sets in terms of hypervolume.
But this correlation also depends on the solution-set size $\mu$.
Let us suppose that the fitness values between neighboring solutions change with a factor~$\alpha$.
Then, the change of the hypervolume values between the corresponding neighboring solution-sets is lower than $\alpha$.
Notice that the magnitude of the autocorrelation length relative to the hypervolume is approximately $\mu$ times
the one related to the solution-level fitness values.
Nevertheless, as the well-known result from single-objective $NK$-landscapes,
the autocorrelation length of the hypervolume decreases 
with the non-linearity degree of $\rho MNK$-landscapes (Figure \ref{fig:autocorLength} -- bottom).
With respect to the objective space dimension and to the objective correlation,
the autocorrelation lengths are nearly the same.

Our results shed new lights on the definition of a moFiL.
According to the hypervolume indicator and to the very elementary neighborhood used in the experiments,
the structure of the moFiL is very smooth.
The ruggedness of the landscapes depends more on the non-linearity than on the objective space dimension or on the objective correlation.
This gives complementary information with respect to \cite{verel2011b},
which enlighten the importance of objective correlation and objective space dimension on the structure of the Pareto optimal set.
Moreover, from the algorithm-design perspective,
if we refer on results from single-objective fitness landscapes analysis,
a local search based on solution-sets and on the hypervolume should be efficient for the $\rho MNK$-landscapes.

\subsection{Adaptive Walk}
\begin{figure*}[p]
\begin{center}
\begin{tabular}{cc}
\includegraphics[width=0.34\textwidth]{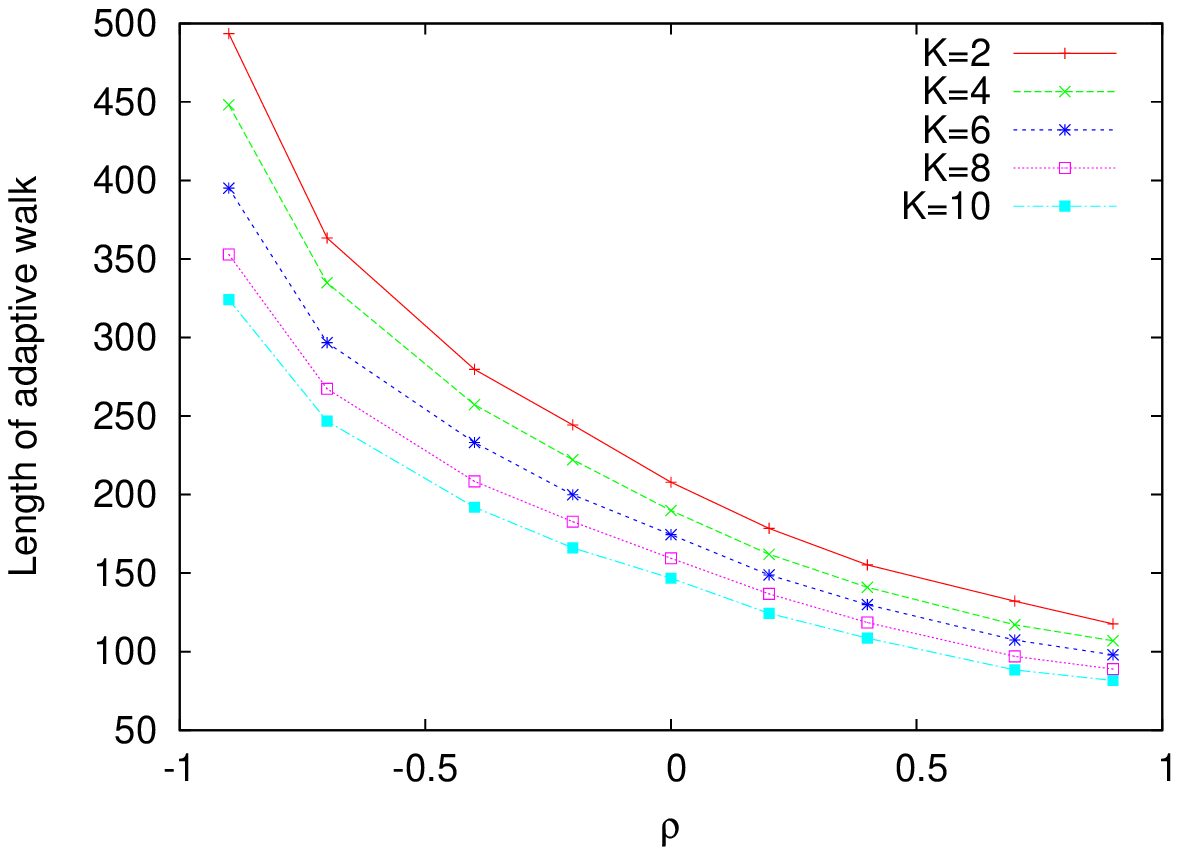} & \includegraphics[width=0.34\textwidth]{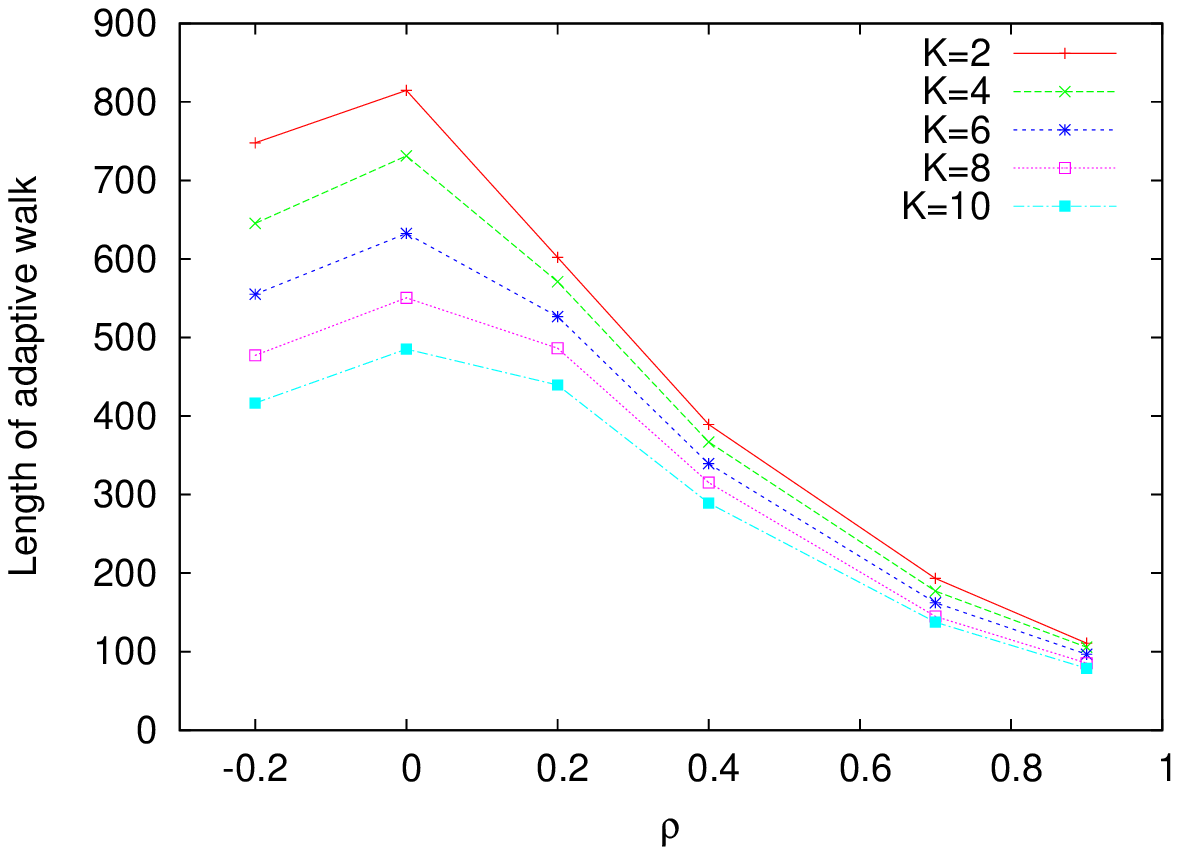} \\
\includegraphics[width=0.34\textwidth]{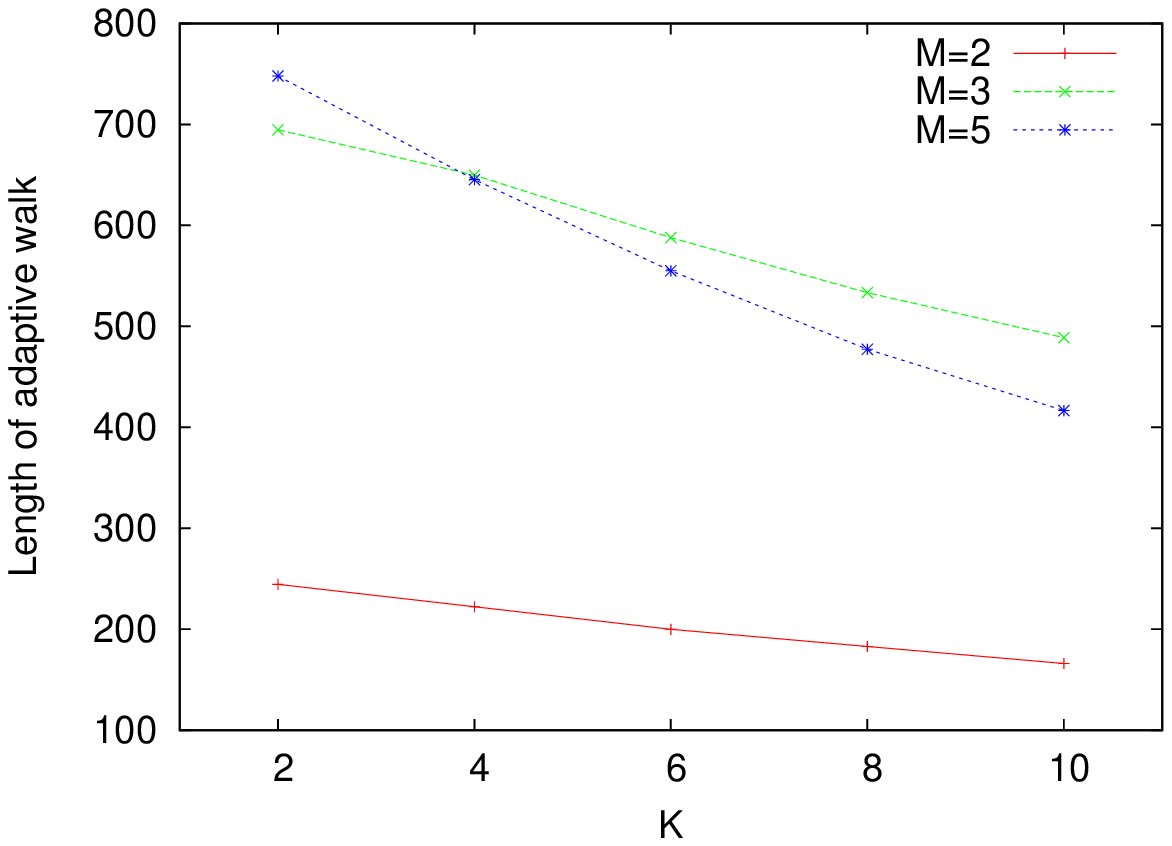} & \includegraphics[width=0.34\textwidth]{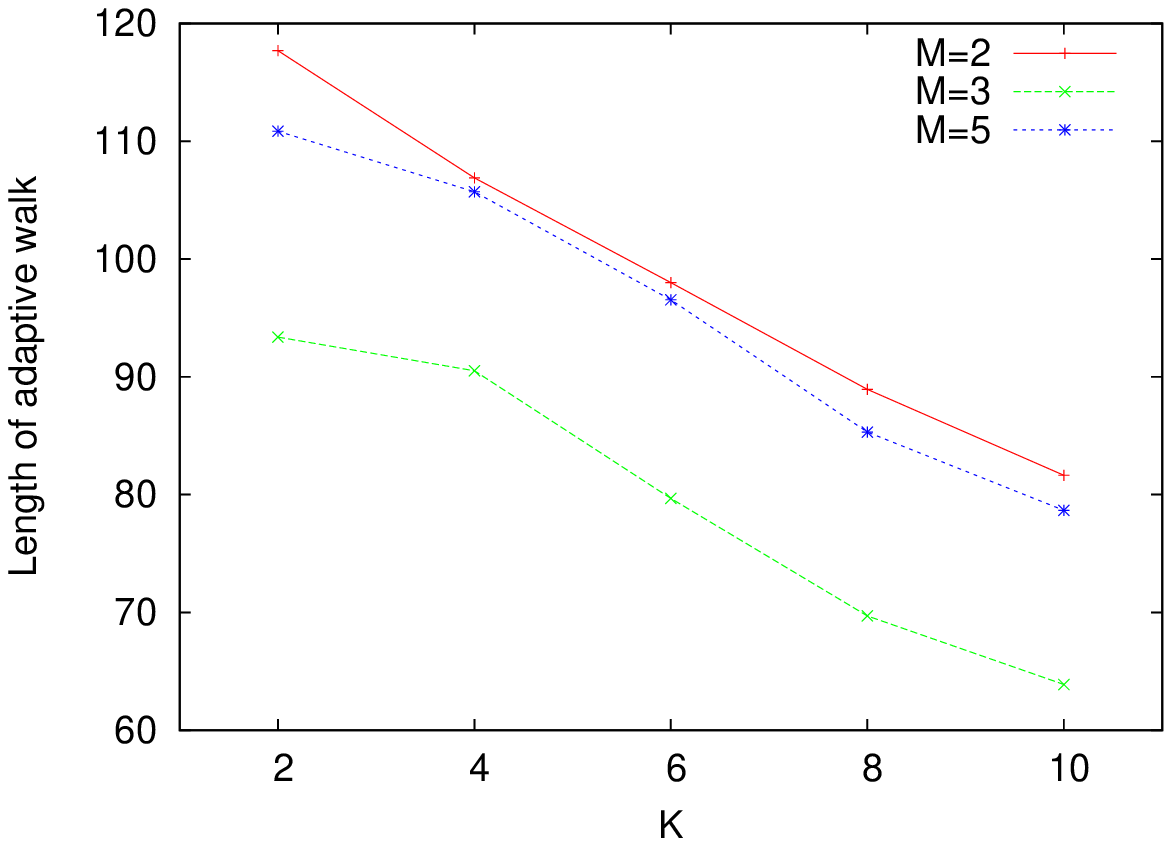} \\
\end{tabular}
\caption{Average length of the adaptive walks
according to parameter $\rho$ (top left $M=2$, right $M=5$),
and to parameter $K$ 
(bottom left $\rho=-0.2$, right $\rho=0.9$).
The solution-set size is $\mu=20$.
\label{fig:adaptiveLength}}
\end{center}
\end{figure*}
%
\begin{figure*}[t]
\begin{center}
\begin{tabular}{cc}
\includegraphics[width=0.34\textwidth]{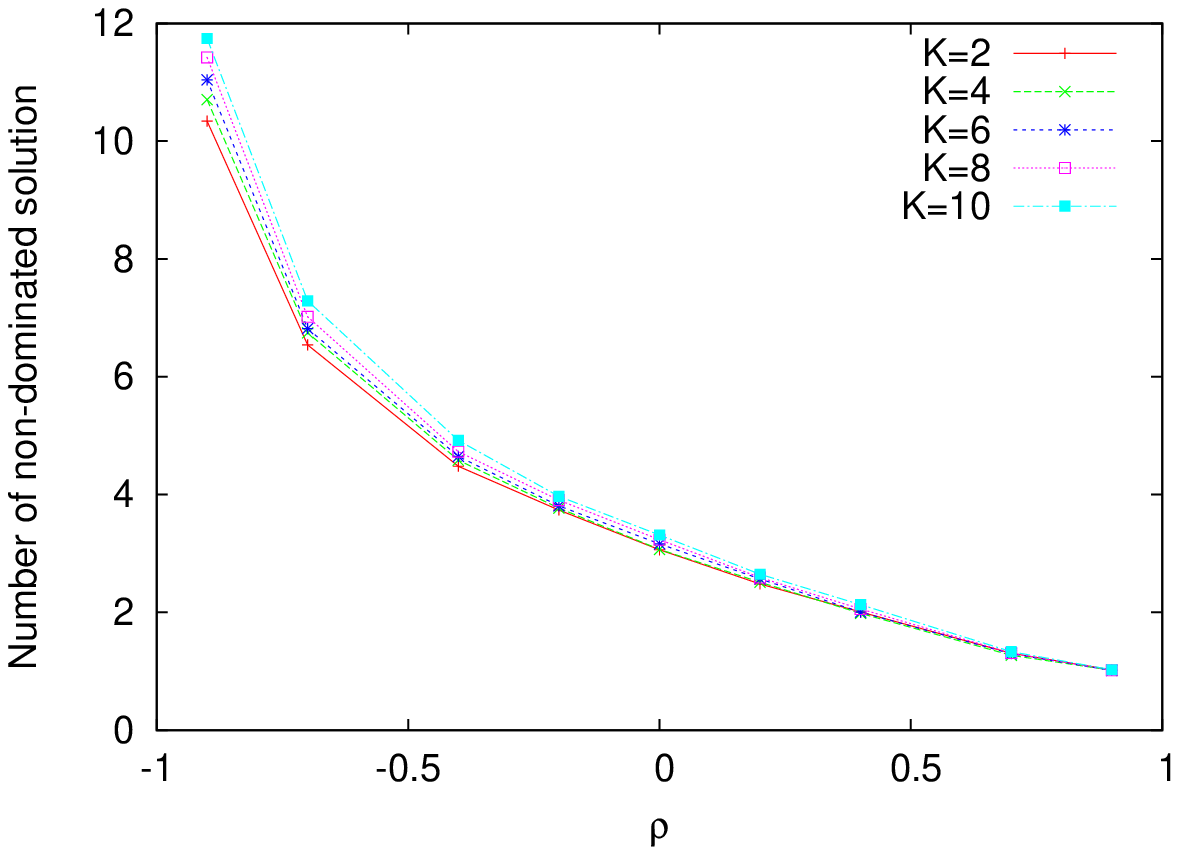} & \includegraphics[width=0.34\textwidth]{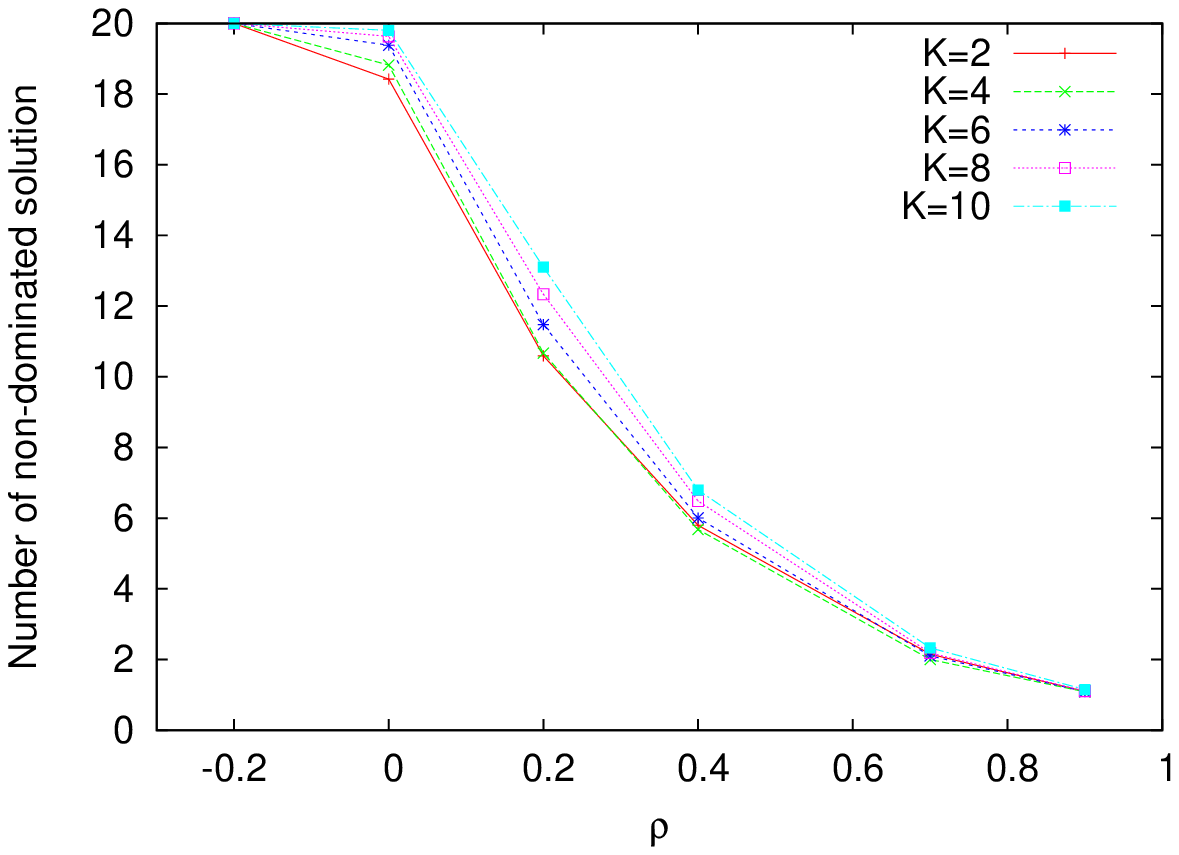} \\
\includegraphics[width=0.34\textwidth]{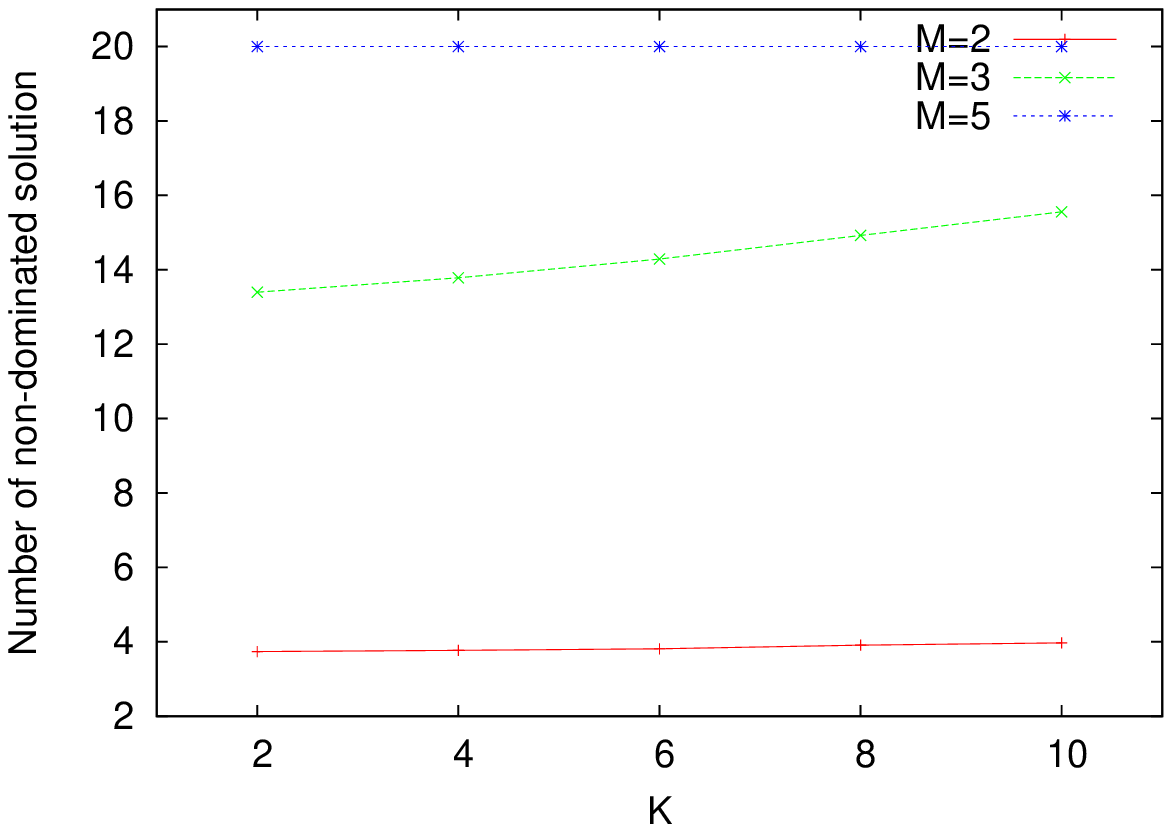} & \includegraphics[width=0.34\textwidth]{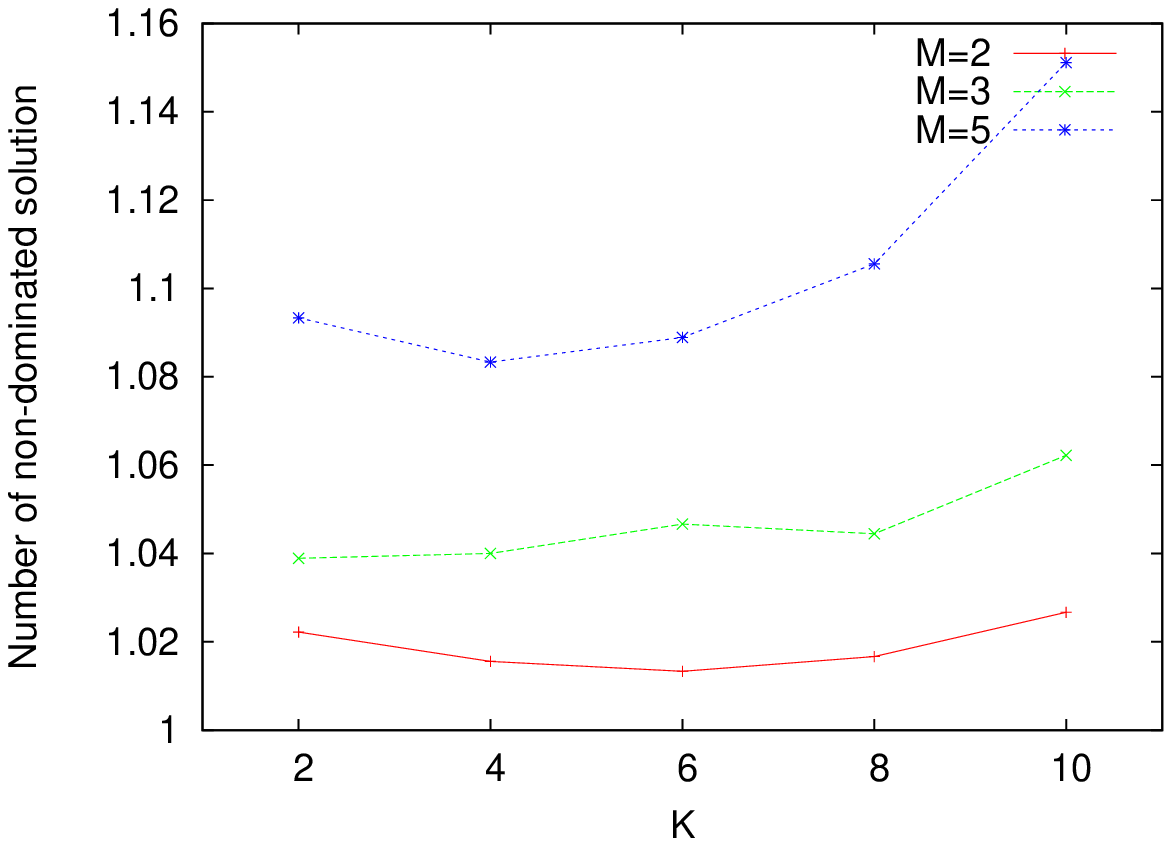} \\
\end{tabular}
\caption{Average number non-dominated solutions in the solution-set local optima 
according to parameter~$\rho$ (top left $M=2$, right $M=5$),
and to parameter $K$ 
(bottom left $\rho=-0.2$, right $\rho=0.9$).
The solution-set size is $\mu=20$.
\label{fig:size}}
\end{center}
\end{figure*}

In this section, we define an adaptive walk as a first-improvement hill-climbing (HC) algorithm performing on solution-sets.
At each algorithm iteration, a random neighboring solution-set is accepted
if its hypervolume-value is strictly better than the one of the current solution-set.
The walk stops once a local optimum solution-set is found, according to the set-level neighborhood relation.
The length of the adaptive walks is studied with a solution-set size $\mu=20$.
It reduces the size of the neighborhood structure and then, of the time complexity of the HC algorithm.
Usually, it is expected that, when the problem difficulty increases, so is the number of local optima.
As a consequence, the length to reach a local optimum becomes smaller.

Figure \ref{fig:adaptiveLength} shows the length of the adaptive walks according to the $\rho MNK$-landscapes parameters.
First, as expected, for a fixed objective space dimension and objective correlation,
the length of adaptive walks decrease with the non-linearity degree $K$.
The length is correlated to the difficulty of the problem under study.
However, surprisingly, the length decreases when the objective correlation increases,
whereas, intuitively, the search becomes easier when the objective correlation increases.
A notable exception stands  for $M=5$, with $\rho \in \{-0.2, 0.0\}$.

In order to explain this result, we need to deeply analyze the set-based HC.
Let us note that only non-dominated solutions from the set contribute to the hypervolume.
As a consequence, when the number of non-dominated solutions is small,
the number of neighboring solution-sets with a strictly higher hypervolume-value is small.
In such a case, the length of the adaptive walk should be smaller.
This should explain our results.
Indeed, according to \cite{verel2011b},
the size of the Pareto optimal set increases when the objective space dimension and the objective correlation decrease.
The non-linearity $K$ has a low influence on this size.
Figure~\ref{fig:size} shows the number of mutually non-dominated solutions in the output of the algorithm
(\IE in the solution-set local optima).
As the size of the Pareto optimal set, the number of non-dominated solutions in the set decreases with the objective correlation,
and is nearly constant with the parameter $K$.
For $M=2$, the maximum size $\mu=20$ is never reached.
For $M=3$, the maximum size is nearly reached for all correlation values between $\rho = -0.2$ and $\rho = 0.0$.

When there is an equivalent number of mutually non-dominated solutions in the solution-sets of the different landscapes,
the length of adaptive walks corroborates the expected property: 
the larger the size, the `easier' the problem.
In such a case, like in single-objective fitness landscapes,
the length of adaptive walks could be used to estimate the diameter of the basins of attraction of local optima.
However, three possible ways could overcome the drawback related to the number of non-dominated solutions. 
First, it is possible to change the search space or the neighborhood relation
in order to consider mutually non-dominated solutions in the sets only.
The indicator-based fitness function could also be modified in order to take dominated solutions into account.
Second, we can change the definition of the HC algorithm in order to consider the ties in hypervolume-values.
At last, when there is a large number of neighboring solution-sets sharing the same hypervolume-value,
we can see the fitness landscapes as covered by many plateaus. 
Then, it could become useful to study the structure of the plateaus more than the solution-set local optima.
The decision between these choices has to be made depending on the issues to analyze,
and according to the problem and the algorithm under consideration.

\begin{figure}
\begin{center}
\includegraphics[width=0.345\textwidth]{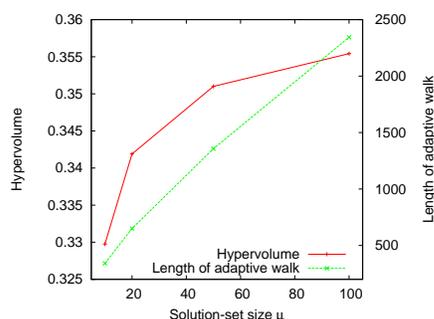}
\caption{
Average hypervolume of the solution-set local optima,
and average length of the adaptive walks according to the solution-set size $\mu$.
The number of objectives is $M=3$, 
the objective correlation is $\rho = -0.2$, 
and the non-linearity degree is $K=4$.  
}
\label{fig:setSize}
\end{center}
\end{figure}

At last, as shown in Figure \ref{fig:setSize}, the size of the solution-sets impacts the length of adaptive walks.
With the cost of additional evaluations, the quality of the set-based local optimum increases with the solution-set size.
Indeed, the length of adaptive walks and the hypervolume-value
 both increase with the solution-set size, but at different rates.
This suggests that a trade-off between cost and quality exists with respect to the solution-set size.

%

\section{Conclusions}

In this paper, we formulated a definition of set-based multiobjective fitness landscapes.
It is based on a set of solution-sets as a search space,
an indicator quantifying the quality of solution-sets as a fitness function,
and a set-based neighborhood relation.
We performed a set-based multiobjective fitness landscape analysis on the multiobjective $NK$-landscapes with objective correlation.
Our preliminary experimental study shows that tools from single-objective fitness landscapes can directly be 
extended for analyzing set-based multiobjective search approaches.
The relevant features of multimodality and ruggedness has been highlighted for this particular class of problems.

Two difficulties have been pointed out in this work.
First, the size of the set-based neighborhood can become very large in comparison with solution-based neighborhood structures.
Second, some solutions contained in a feasible solution-set may become dominated by others,
so that they do not contribute to indicator-based fitness values for most existing quality indicators.
As a consequence, future methodologies will be devoted to an efficient way of sampling the neighborhood,
while taking dominated solutions into account.
As a next step, we will formalize existing multiobjective search algorithms in terms of set problems based on a set-based neighborhood. 
Such advances will allow us to analyze the link between the performance and the dynamics of given search methods,
together with the main features of multiobjective fitness landscapes.
Moreover, we plan to experiment more advance concepts, related to the evolvability, the neutrality, or local optima networks
in order to enlarge the understanding of multiobjective problem structures.

\bibliographystyle{abbrv}

\end{document}